\documentclass[conf]{new-aiaa}
\usepackage[utf8]{inputenc}

\usepackage{graphicx}
\usepackage{amsmath}
\usepackage[version=4]{mhchem}
\usepackage{siunitx}
\usepackage{longtable,tabularx}
\usepackage{float}
\usepackage{subcaption}

\usepackage{color}  
\RequirePackage[dvipsnames,usenames]{xcolor}

\title{Upgrading from Gaussian Processes to Student's-T Processes}

\author{Brendan D. Tracey \footnote{Postdoctoral Researcher, tracey.brendan@gmail.com}}
\affil{Santa Fe Institute, Santa Fe, NM, 87501 \\
Massachusetts Institute of Technology, Cambridge, MA, 02139}
\author{David H. Wolpert \footnote{Resident Faculty, dhw@santafe.edu}}
\affil{Santa Fe Institute, Santa Fe, NM, 87501 \\
Massachusetts Institute of Technology, Cambridge, MA, 02139 \\
Arizona State University, Tempe, AZ, 85287}

\begin{document}
\maketitle

\begin{abstract}

Gaussian process priors are commonly used in aerospace design for performing Bayesian optimization. Nonetheless, Gaussian processes suffer two significant drawbacks: outliers are \emph{a priori} assumed unlikely, and the posterior variance conditioned on observed data 
depends only on the locations of those data, not the associated sample values.
Student's-T processes are a generalization of Gaussian processes, founded on the Student's-T distribution instead of the Gaussian distribution. Student's-T processes maintain the primary advantages of Gaussian processes (kernel function, analytic update rule) with additional benefits beyond Gaussian processes. The Student's-T distribution has higher Kurtosis than a Gaussian distribution and so outliers are much more likely, and the posterior variance increases or decreases depending on the variance of observed data sample values.
Here, we describe Student's-T processes, and discuss their advantages in the context of aerospace optimization. 
We show how to construct a Student's-T process using a kernel function and how to update the process given new samples. We provide a clear derivation of optimization-relevant quantities such as expected improvement, and contrast with the related computations for Gaussian processes. Finally, we compare the performance of Student's-T processes against Gaussian process on canonical test problems in Bayesian optimization, and apply the Student's-T process to the optimization of an aerostructural design problem.

\end{abstract}

\section{Nomenclature}

{\renewcommand\arraystretch{1.0}
\noindent\begin{longtable*}{@{}l @{\quad=\quad} l@{}}

$D$ & Data set of inputs and observed function values at those inputs \\
$\mathbb{E}[x]$ & Expected value of random variable $x$ \\
$EI$ & Expected Improvement \\
$f$ & Objective function of interest \\
$GP$ & Gaussian process \\
$K$ & Matrix of pairwise kernel function evaluations \\ 
$k(x, x')$ & Kernel function of a process \\
$m(x)$ & Mean function of a process \\
$MVG$ & Multivariate Gaussian distribution \\
$MVT$ & Multivariate Student's-T distribution \\
$p(y|x)$ & Marginal probability of $y$ at an input $x$ \\
$p(y|x, D)$ & Marginal probability of $y$ at an input $x$ given data $D$\\
$STP$ & Student's-T process \\
$\mathcal{T}$ & Multivariate Student's-T distribution \\
$x$ & Input to the objective function $f$ \\
$\tilde{x}$ & Observed input to the objective function $f$ (with corresponding $\tilde{y}$) \\
$y$ & Output from the objective $y$ \\
$y^*$ & Global optimum of a function \\
$\hat{y}$ & Best current known function value \\
$\tilde{y}$ & Observed output from the objective $y$ \\
$\mu$ & Mean parameter of the distribution \\
$\nu$ & Degrees of freedom parameter for the Student's-T distribution/process \\
$\Sigma$ & Shape parameter of the distribution \\
$\sigma$ & Marginal shape parameter at a given input \\
$\Phi$ & Cumulative distribution function of a standard Gaussian distribution \\
$\Phi_s$ & Cumulative distribution function of a standard Student's-T distribution \\
$\phi$ & Probability density function of a standard Gaussian distribution \\
$\phi_s$ & Probability density function of a standard Student's-T distribution \\
$\xi$ & Constraint violation

\end{longtable*}}

\section{Introduction}
One major challenge in aerospace design optimization is the computational expense in determining the quantity of interest.
Even an Euler flow solution around a full aircraft geometry is still a non-trivial endeavor, let alone computing the aerodynamic forces with higher-fidelity governing equations like RANS and DES. Nonetheless,
these higher fidelity simulations are being increasingly relied upon by the design community, a trend which will only increase as researchers work to reduce the current deficiencies in computational fluid dynamics (CFD), such as turbulent transition, separation, and engine stall analysis \cite{slotnick2014cfd}. There is accordingly a premium on 
sampling (data-generation) efficiency when designing optimization algorithms when
the objective function is such a computationally expensive simulation.

Bayesian optimization is a class of techniques with several advantages in finding global optima
of objective functions that are computationally expensive to sample (e.g., high-fidelity
simulators). Perhaps their primary advantage is that they need relatively
few samples to find global minima of a function, in contrast to local hill-descending methods such as the Nelder-Mead simplex \cite{nelder1965simplex} or quasi-Newton methods \cite{nocedal2006numerical} that only find local minima, and in contrast to population- or sampling-based global methods such as genetic algorithms \cite{deb2000fast} or simulated annealing \cite{van1987simulated} that typically require huge sample sets 
(e.g. hundreds or thousands of simulations in an aerospace design context).
Bayesian optimization techniques work by defining a relative probability of different function behaviors in the form of a probabilistic \emph{prior} over the space of functions.
As the objective function is repeatedly evaluated, this prior distribution can be updated using Bayes rule to get an updated belief, i.e. \emph{posterior}, about the function behavior at unobserved inputs. This Bayesian update is often computationally expensive relative to other classes of optimization algorithms, but this added expense is often negligible relative to (for
example) the sampling cost of evaluating a flow simulation.

The most common prior used in Bayesian optimization is a Gaussian process (GP). A Gaussian process assumes that the marginal joint distribution of function values at any (finite) set of input locations is a multivariate Gaussian distribution (MVG). Gaussian processes have several desirable mathematical properties. In particular, the Bayesian update step is analytic, as is finding the marginal distributions for the function behavior at unknown locations. Gaussian processes have been widely used in global optimization, from the important EGO algorithm \cite{jones1998efficient}, to supersonic jet optimization \cite{lukaczyk2013managing}, to multi-fidelity modeling and optimization \cite{lam2015multifidelity}.  

In spite of their success, GPs have several known shortcomings. First, the Gaussian distribution is not a ``heavy-tailed'' distribution, and so the Bayesian posterior is forced to assign
low probability to extreme outliers --- regardless of the data.
Second, the posterior variance of a Gaussian process does not depend on the returned objective values, but depends \emph{only} on the input evaluation locations. This implies, for instance, that the posterior variance is not higher if the objective values in the sample
set are all much different than expected under the Gaussian prior. One way to deal with this is to define a hyperprior  over GP parameters (such as the observation noise), but then evaluating the posterior is often not a simple update equation, instead requiring approximate inference algorithms such as Markov chain Monte Carlo.

In this paper, we argue for the use of a different probabilistic prior ---  a Student's-T process (STP). Student's-T processes assume that the function values at any (finite) set of input locations are jointly distributed according to a multivariate Student's-T distribution (MVT), unlike the MVG of a Gaussian process. Like a GP, there is an analytic formula for updating an STP with new samples, and it is easy to find the marginal distribution for unknown locations. Additionally, the Student's-T distribution includes an additional \emph{degrees of freedom} parameter controlling the kurtosis of the distribution. This means outliers can be much more likely under an STP than a GP, In addition, as will be shown, the posterior variance is increased if observed values vary more than expected under the corresponding Gaussian process, and the posterior variance is decreased if they vary by less than expected. As the degrees of freedom parameter approaches infinity the STP converges to a GP, and so STPs are a generalization of GPs.  These properties make STPs well-suited for aerospace optimization problems. Aerospace design problems often feature smooth regions punctuated by (near) discontinuities, for example the transition to stall, or the failure to meet constraints. These discontinuities would be considered extremely unlikely according to a Gaussian prior, but not so under a Student's-T prior. 

The paper is organized as follows. We begin by reviewing Gaussian processes. We then introduce the Student's-T process, comparing and contrasting it with GPs. We then derive formulae useful for Bayesian optimization, such as expected improvement and marginal likelihood. Finally, we compare the performance of GPs with STPs on benchmark optimization problems and an aerostructural design optimization problem. We find that the Student's-T processes significantly outperform their Gaussian counterparts. In the future, this work can be extended to more complex design examples, for example with function gradients and constraints, and to other Bayesian search processes.

\section{Gaussian Process Review}
In this section we briefly review Gaussian processes in a way
that facilitates comparison with Student's-T processes, which we describe in the 
following section. A more complete description of Gaussian processes can be found in \cite{rasmussen2006gaussian}.


A GP is a ``collection of random variables, any finite number of which have a joint Gaussian distribution''\cite{rasmussen2006gaussian}. A GP gives a prior over the space of possible
objective functions, parameterized by two functions. The first is a mean function, $m(x)$, that sets the prior expected value of the objective function at every input location $x$. The second is a kernel function $k(x,x')$, that sets the covariance between the values
of the objective function at any two input locations $x$ and $x'$. 

We will write the GP prior distribution over objective functions specified by such a set
of parameters as
\begin{equation}
f(x) \sim \mathcal{GP}(m(x), k(x,x'))
\end{equation}
At a single input location, $x$, the prior distribution of
the objective function is described by a (univariate) Gaussian distribution
\begin{align}
p(y|x) = \mathcal{N}(\mu, \sigma)(y)
\end{align}
where $\mu = m(x)$, and $\sigma = k(x,x)$. At multiple input locations, the joint marginal prior distribution is described by a multivariate Gaussian
\begin{align}
p(y|x) = \mathcal{N}(\mu, \Sigma)(y)
\end{align}
where $\mu_i = m(x_i)$, and $\Sigma_{i,j} = k(x_i, x_j)$.

It is often the case that $m(x)$ is assumed to be 0, and this will be assumed for the remainder of the work.
The kernel function is typically chosen so that the covariance decreases as $x$ and $x'$ become farther apart.  A common choice is the ``isotropic squared exponential'' function (SE)
\begin{equation}
\label{eq:kerisose}
k(x,x') = exp(-||x-x'||_2^2 \; / \; 2\sigma^2)
\end{equation}
where $\sigma$ is a \emph{bandwidth} hyperparameter that sets the correlation length.

As in standard Bayesian analysis,
this prior can be updated every time the objective function is evaluated at a given $x$ to produce
some value $y$. An important property of the Gaussian distribution is that it is closed under such conditioning on new data, i.e. the posterior distribution after observing a set of samples is still a Gaussian distribution in the other variables. Specifically, it can be shown that given a set of inputs and outputs $D = \{\{\tilde{x_1}, \tilde{y_1}\}, \{\tilde{x_2}, \tilde{y_2}\}, \{\tilde{x_3}, \tilde{y_3}\}, ... \}$, the distribution of values of the objective function $y$ at unobserved locations $x$ is given by
\begin{align}
\label{eq:gppy}
p(y|x,D) &= \mathcal{N}(\mu, \Sigma)(y) \\
\label{eq:gpmu}
\mu &= K_{x,\tilde{x}}, K^{-1}_{\tilde{x},\tilde{x}} \tilde{y} \\
\label{eq:gpsigma}
\Sigma &= K_{x ,x} -  K_{x,\tilde{x}} K_{\tilde{x},\tilde{x}}^{-1} K_{\tilde{x},x}
\end{align}
where $K_{\tilde{x},\tilde{x}}$ is the covariance matrix defined by the kernel function between the observed locations in $D$, $K_{\tilde{x},\tilde{x}}$ is the covariance between the observed locations and the unobserved locations, and $K_{x,x}$ is the covariance among the unobserved locations. It is notable that the posterior covariance matrix \eqref{eq:gpsigma} \emph{only} depends on the observed locations, $\tilde{x}$, and not on the function values themselves $\tilde{y}$. 

GPs are often used for function minimization, where the goal is to find an input location 
which if sampled is likely to have a relatively low value of the objective function. There are many strategies for choosing the next input to sample, such as entropy search \cite{hennig2012entropy} or knowledge gradient \cite{frazier2009knowledge}. The most common strategy, and one of the simplest, is to select the input with the highest \emph{expected improvement} beyond the current best known value. This
means that given a current best function value, $\hat{y}$, the next $x$ to sample is 
\begin{equation}
    \label{eq:gpeiargmax}
argmax_x~EI(x) = argmax_x \int 1\{y < \hat{y}\} (\hat{y}-y) p(y|x,D)dy
\end{equation}
where $p(y|x,D)$ is computed according to \eqref{eq:gppy}, and $1\{.\}$ is the
indicator function that returns 1 if its argument is true and 0 otherwise. The expected improvement under a GP has an analytic form
\begin{equation}
\label{eq:gpei}
EI(x) = (\hat{y} - \mu) \Phi(z) + \sigma * \phi(z)
\end{equation}
where $z = (\hat{y}-\mu)/\sigma$, and $\Phi$ and $\phi$ are the cumulative density function (CDF) and probability density (PDF) for a standard Gaussian distribution respectively. This analytic expression for expected improvement makes it relatively efficient to solve the optimization problem in \eqref{eq:gpeiargmax}.

\section{Student's-T Processes}

Student's-T processes are an alternate prior for Bayesian optimization. Just as GPs have marginal distributions described by the multivariate Gaussian distribution, STPs have marginal distributions described by the multivariate Student's-T distribution \cite{genz2009computation}. 
Student's-T processes receive mention in \cite{rasmussen2006gaussian} and have been used occasionally in modeling \cite{yu2007robust, archambeau2011multiple}. However, a renewal of interest began with Shah et. al. \cite{shah2014student} in which the authors derive the Student's-T process from a Wishart prior, and show that STPs are the most general elliptic process  with analytic density (loosely, elliptic distributions are a class of distributions that are unimodal and where the likelihood decreases as the distance from the mode increases). The process has been further explored for Bayesian optimization \cite{shahbayesian}.

The multivariate Student's-T distribution is a generalization of the multivariate Gaussian distribution with an additional parameter, $\nu$, describing the degrees of freedom of the distribution.
The probability density is given by:
\begin{align}
\label{eq:mvt}
\mathcal{T}(\mu, \Sigma,\nu)(y) = \frac{\Gamma((\nu+d)/2)}{\Gamma(\nu/2)\nu^{d/2}\pi^{d/2} |\Sigma|^{1/2}} \left(1+\frac{1}{\nu} (y-\mu)^T \Sigma^{-1} (y-\mu)\right)^{-(\nu+d)/2}
\end{align}
where $d$ is the dimension of the distribution, $\mu$ is a location parameter, $\Sigma$ is a symmetric positive definite shape parameter, and $\nu > 2$ is the degrees of freedom. Like in a Gaussian distribution, $\mu$ is the mean (and mode) of the distribution.
The shape parameter $\Sigma$ is not the covariance matrix of the distribution, but is related to it,
\begin{equation}
E[(y-\mu)^T(y-\mu)] = \frac{\nu}{\nu -2} \Sigma
\end{equation}
As the degrees of freedom increases, i.e. $\nu \rightarrow \infty$, the multivariate Student's-T distribution converges to a multivariate Gaussian distribution with the same mean and shape parameter.

Like the GP, the Student's-T process is parameterized by a mean function and a kernel function.
However it has one additional parameter, the degrees of freedom, $\nu$. We will write 
the STP prior distribution over objective functions specified by such a set
of parameters as
\begin{equation}
f(x) \sim \mathcal{STP}(m(x), k(x,x'),\nu)
\end{equation}
As in a GP,
the mean function $m(x)$ defines the prior expected value at each location, $x$,
and the kernel function sets the \emph{covariance} between values of
the objective function at any two locations $x$ and $x'$. The joint distribution for a finite subset of locations is
\begin{align}
p(y|x) = \mathcal{T} \left( \mu,\Sigma, \nu \right)(y) =  \mathcal{T} \left( \mu,\frac{\nu -2}{\nu} K, \nu \right)(y)
\end{align}
where $\mu$ is the vector of means, $\mu_i = m(x_i)$, and $K$ is the matrix of pairwise-kernel evaluations $K_{i,j} = k(x_i, x_j)$ (remembering the covariance is not the same as the shape parameter). 

The multivariate Student's-T distribution, and by extension the Student's-T process, is closed under conditioning. Specifically, it can be shown \cite{shah2014student, roth2012multivariate} that given a set of samples, $D = \{\{\tilde{x_1}, \tilde{y_1}\}, \{\tilde{x_2}, \tilde{y_2}\}, \{\tilde{x_3}, \tilde{y_3}\}, ... \}$, the posterior distribution is given by
\begin{align}
\label{eq:tpy}
p(y|x,D) &= \mathcal{T}(\hat{\mu}, \frac{\nu-2}{\nu} \hat{K},\hat{\nu})(y) \\
\label{eq:tmu}
\hat{\mu} &= K_{x,x} K^{-1}_{\tilde{x},\tilde{x}} \tilde{y} \\
\label{eq:tsigma}
\hat{K} &= \frac{\nu + \tilde{y}^T K_{\tilde{x},\tilde{x}}^{-1} \tilde{y}-2}{\nu + |D|-2} \left(
 K_{x,x} -  K_{x,\tilde{x}} K_{\tilde{x},\tilde{x}}^{-1} K_{\tilde{x},x} \right) \\
\hat{\nu} &= \nu + |D|
\end{align}
Comparing \eqref{eq:gpmu} and \eqref{eq:tmu}, we see that
the posterior mean for an STP is identical to that of a GP (for the same kernel function).
The posterior covariance, however, is different between the two processes. 
The rightmost term in \eqref{eq:tsigma} matches its Gaussian counterpart, but the leading term has no equivalent in a GP. This term has an explicit dependence on $\tilde{y}$, the function outputs, and scales the posterior covariance.
If $\tilde{y}^T K_{\tilde{x},\tilde{x}}^{-1} \tilde{y}$ is greater than $|D|$, then the posterior covariance is larger than its Gaussian counterpart, while if
it is less, the posterior covariance is lower. 

When do these conditions hold?
It can be shown that the squared Mahalanobis distance for Gaussian-generated samples is distributed according to a $\chi^2$ distribution with a mean $|D|$ \cite{slotani1964tolerance}. This means $\tilde{y}^T K_{\tilde{x},\tilde{x}}^{-1} \tilde{y} \approx |D|$ (in expectation) \emph{if} the output values are actually generated from a Gaussian processes, i.e. $y \sim \mathcal{N}(0,\Sigma)$. Similarly, this implies that if the observed $y$ values vary from each other by about as much as one would expect under a GP, then the posterior covariance under an STP is roughly identical to the covariance of the equivalent GP. 
 On the other hand, if the values vary by significantly more or less than expected, then the posterior uncertainty under the STP is significantly higher or lower. Note that as, $\nu \rightarrow \infty$, 
both $\tilde{y}^T K_{\tilde{x},\tilde{x}}^{-1} \tilde{y} \rightarrow |D|$ and the difference between the terms matters less. So the difference between the STP and GP predictions is most prominent for small values of $\nu$. In the small-$\nu$ regime, however, an STP will be more adaptive to the observed samples at the cost of regularization provided by the Gaussian assumption.

The benefits of using an STP do not come with a significantly higher computational cost. The only major computational difference between the two update rules is the additional term in \eqref{eq:tsigma}. However, for both processes, the dominant cost is computing the Cholesky decomposition of $K_{\tilde{x},\tilde{x}}$ to compute terms involving $K_{\tilde{x},\tilde{x}}^{-1}$, which can be computed once and cached. In fact, the expression $K_{\tilde{x},\tilde{x}}^{-1} \tilde{y}$ occurs in both \eqref{eq:tmu} and \eqref{eq:tsigma}, so if the mean prediction has already been computed, computing the additional scaling factor of the STP only requires computing a dot product, which is $O(|D|)$.

\subsection{Expected Improvement}

Under an STP, the marginal distribution for the output at a single (unobserved) input is a 
Student's-T distribution, as discussed above:
\begin{equation}
\mathcal{T}(\mu,\sigma,\nu) =  \frac{\Gamma(\frac{\nu+1}{2})}{\sqrt{\nu \pi} \Gamma(\frac{\nu}{2})} \frac{1}{\sigma}  \left( 1 + \frac{((y-\mu)/\sigma)^2}{\nu} \right)^{-\frac{\nu+1}{2}}
\end{equation}
Below 
we show that the Student's-T distribution also has an analytic expression for the expected improvement over a given $\hat{y}$.
This implies that selecting the next design location using expected improvement is just
as easy under an STP as under a GP.

To begin, for simplicity, define
\begin{equation}
C = \frac{\Gamma(\frac{\nu+1}{2})}{\sqrt{\nu \pi} \Gamma(\frac{\nu}{2})}
\end{equation}
The expected improvement equation becomes
\begin{align}
\int_{-\infty}^{\infty} 1\{y < \hat{y}\} (\hat{y}-y) p(y|x)dy = \int_{-\infty}^{\hat{y}} (\hat{y}-y) \frac{1}{\sigma} C \left( 1 + \frac{((y-\mu)/\sigma)^2}{\nu} \right)^{-\frac{\nu+1}{2}} dy
\end{align}
First, substitute $t = \frac{y-\mu}{\sigma}$, $dt = dy/\sigma$.
\begin{align}
&= \int_{-\infty}^{\frac{\hat{y}-\mu}{\sigma}} (\hat{y}- \mu -\sigma t) C \left(1 + \frac{t^2}{\nu} \right)^{-\frac{\nu+1}{2}} dt \\
&= (\hat{y}- \mu) \int_{-\infty}^{\frac{\hat{y}-\mu}{\sigma}} C \left(1 + \frac{t^2}{\nu} \right)^{-\frac{\nu+1}{2}} dt  - \int_{-\infty}^{\frac{\hat{y}-\mu}{\sigma}} \sigma t C \left(1 + \frac{t^2}{\nu} \right)^{-\frac{\nu+1}{2}} dt
\end{align}
The integrand of the first integral 
on the RHS is the standard Student's-T distribution (i.e.,
a Student's-T distribution with $\mu = 0$ and $\sigma = 1$). So this integral is just the CDF of the standard T distribution. Next, make the substitution $z = (\hat{y}-\mu)/\sigma$,
and in the second integral on the RHS 
make the substitution $s = 1 + t^2/\nu$, so $t dt = (\nu/2) ds$.
\begin{align}
&= (\hat{y} - \mu) \Phi_s(z) - \sigma \int_{-\infty}^{1+\frac{z^2}{2}} \frac{\nu}{2} C s^{- \frac{\nu + 1}{2}} ds \\
&= (\hat{y} - \mu) \Phi_s(z) - \sigma \frac{\nu}{2} \frac{2}{1-\nu} C s^{-\frac{\nu+1}{2}}s |_{-\infty}^{1+\frac{z^2}{2}} \\
&= (\hat{y} - \mu) \Phi_s(z) + \sigma \frac{\nu}{\nu-1} \left(1+\frac{z^2}{\nu} \right) C \left(1+\frac{z^2}{\nu} \right)^{-\frac{\nu+1}{2}}\\
\label{eq:finaltei}
&= (\hat{y} - \mu) \Phi_s(z) + \frac{\nu}{\nu-1} \left(1+\frac{z^2}{\nu} \right) \sigma \phi_s(z)
\end{align}
The formula \eqref{eq:finaltei} is the expected improvement over $\hat{y}$ for a Student's-T distributed random variable, where again $z = (\hat{y}-\mu)/\sigma$ and $\Phi_s$ and $\phi_s$ are the CDF and PDF of the standard Student's-T distribution respectively ($\mathcal{T}(0,1,\nu)$).  As expected, this converges to \eqref{eq:gpei} as $\nu \rightarrow \infty$. When $\nu$ is small, however, the expected improvement can be significantly different from a Gaussian, even when both distributions have the same mean and standard deviation. A smaller $\nu$ increases the likelihood of outliers, and in comparison to EI under a GP,
more importance is placed on having a large uncertainty than on having a promising mean. The best value of $\nu$ will depend on how to the objective function is likely to behave. If large deviations and outliers are actually expected, then small values of $\nu$ will correctly encourage exploration near existing samples, while if these deviations are unlikely, then a GP (or a larger value of $\nu$) will be better suited for modeling, as it will encourage more exploration.

\subsection{Illustrative Comparison}
\begin{figure*}[t!]
\centering
   \begin{subfigure}[t]{0.45\textwidth}
   \includegraphics[width=1\linewidth]{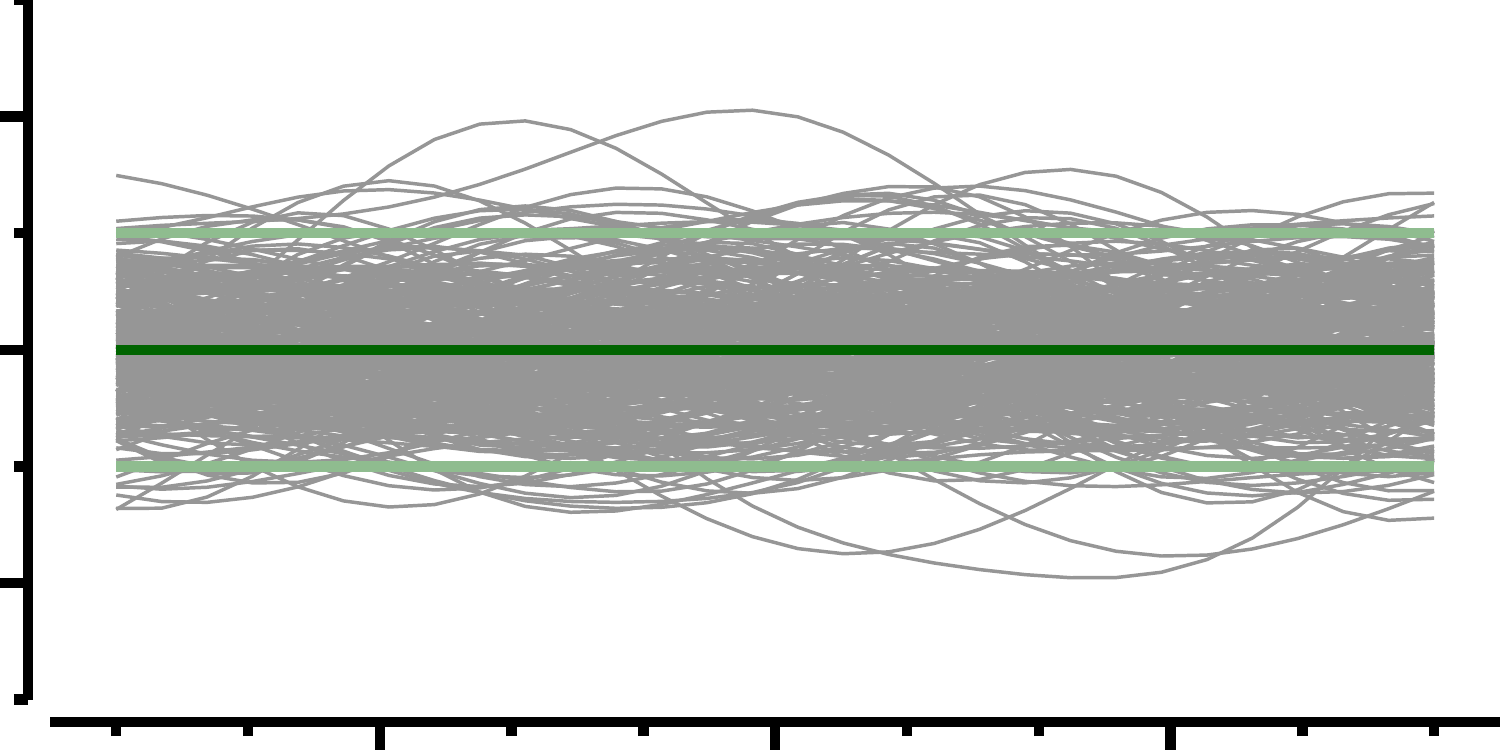}
   \caption{}
   \label{fig:gpnodata}
\end{subfigure}
~
\begin{subfigure}[t]{0.45\textwidth}
   \includegraphics[width=1\linewidth]{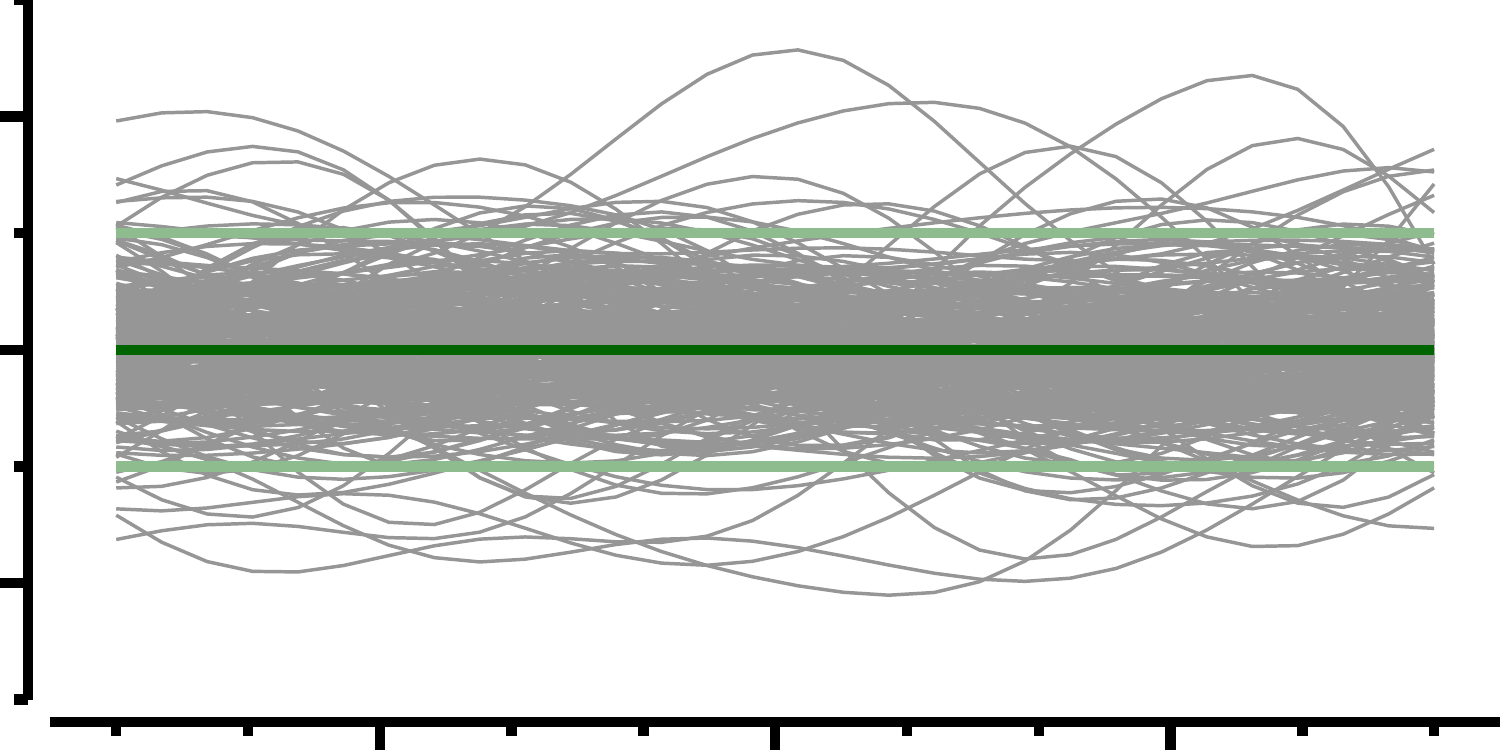}
   \caption{}
   \label{fig:tnodata}
\end{subfigure}
\caption[]{Comparison of  (a) a Gaussian process and (b) a Student's-T process with the same prior mean and kernel function. Notice that extreme outliers are much more likely under the Student's-T process than the Gaussian process.}
\label{fig:priorcomp}
\end{figure*}

We illustrate the difference between the GP and STP processes. First, we compare draws from the prior distribution (before the function has been sampled anywhere). We set both processes to have a zero mean function ($m(x) = 0$), and to use the isotropic squared exponential kernel function, \eqref{eq:kerisose}. The STP is set with $\nu = 5.0$. Fig. \ref{fig:priorcomp} compares realizations of these two processes. The dark green line is the mean of the process at each (1-D) input, and the lighter green lines show the mean plus or minus two standard deviations. The lighter gray lines depict $300$ realizations of the particular process, with Fig. \ref{fig:gpnodata} representing draws from the GP, and Fig. \ref{fig:tnodata} showing realizations from the STP. Despite the fact that the two processes have the same mean and covariance, outliers are significantly more likely under the STP prior. The draws from the Gaussian process are mostly contained within the $2$-standard-deviation error lines, and the most extreme deviations are not much outside it. The STP, on the other hand, has many more outliers, and there are several draws that far exceed the bounds, which would be very unlikely under a Gaussian prior.

\begin{figure}
\centering
   \begin{subfigure}[b]{0.45\textwidth}
   \includegraphics[width=1\linewidth]{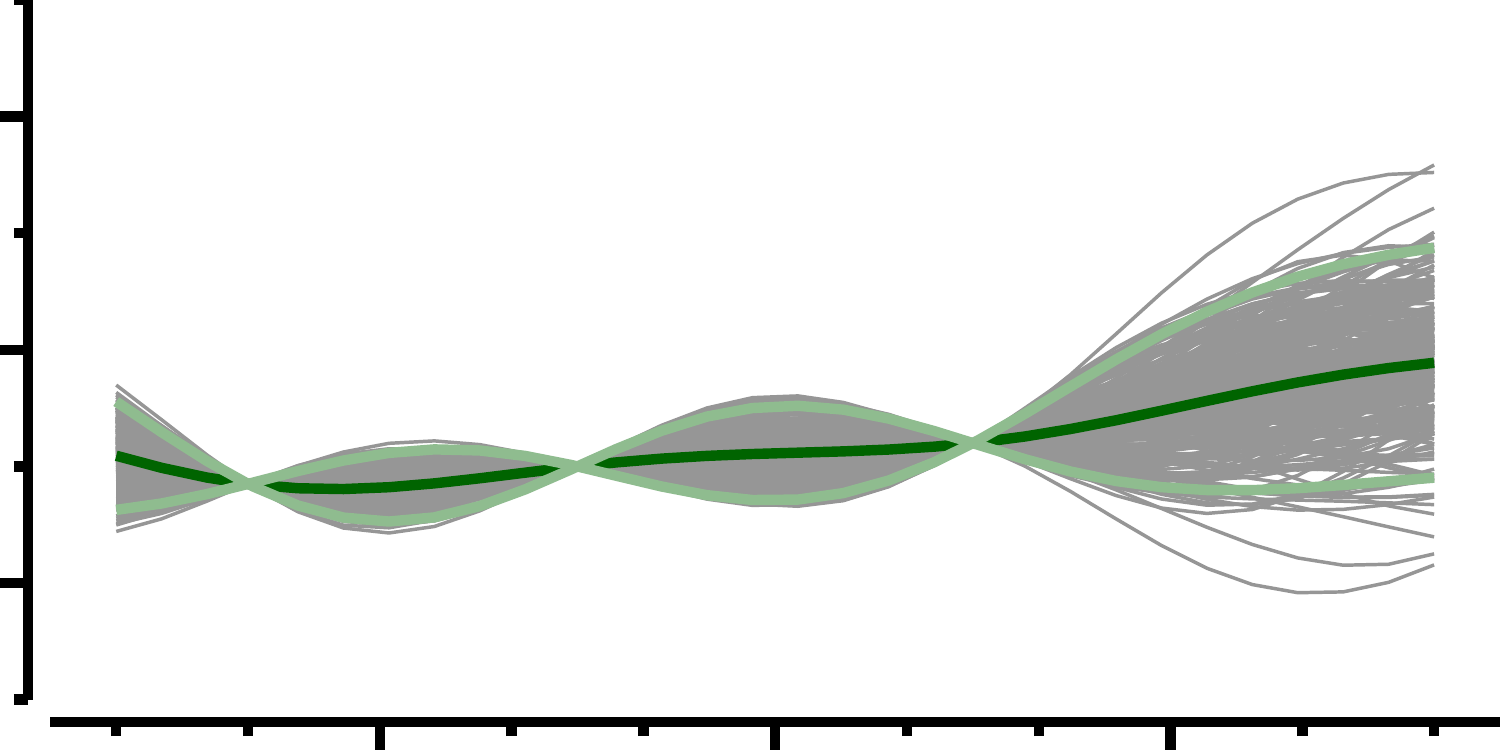}
   \caption{}
   \label{fig:gpdata}
\end{subfigure}
\begin{subfigure}[b]{0.45\textwidth}
   \includegraphics[width=1\linewidth]{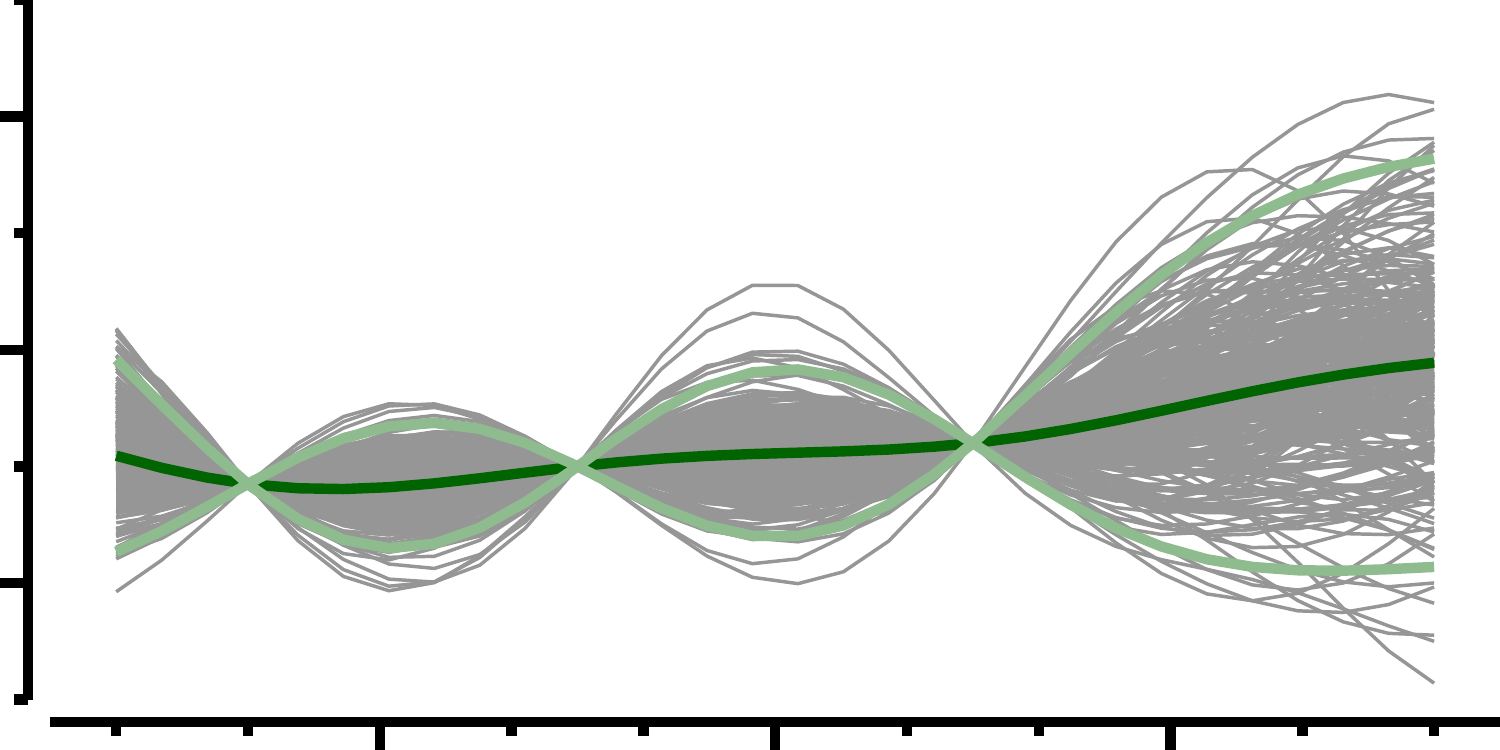}
   \caption{}
   \label{fig:tdata}
\end{subfigure}
\caption[]{Comparison of the posterior for (a) a Gaussian process and (b) a Student's-T process. The two processes start with the same prior mean and covariance and are updated with the same sample data. The posterior uncertainty of the Student's-T distribution is larger than the uncertainty in a GP, which compounded with STPs having larger outliers in general.}
\label{fig:updatedata}
\end{figure}

We next compare the processes after observing data. We update the GP and STP priors using the same set of input/output data   (synthetically created for demonstration purposes) 
using the formulae in \eqref{eq:gppy} and \eqref{eq:tpy} respectively.
The results of this update are shown in Fig. \ref{fig:updatedata}, with Fig. \ref{fig:gpdata} showing the posterior of the GP, and Fig. \ref{fig:tdata} showing the posterior of the STP.
We first observe that the STP is more likely to generate large outliers, just as under the prior. The GP only has significant outliers near the edge of the domain, while the STP has significant outliers not only on the edge of the domain, but also within the interior. Second, we see that the posterior uncertainty is generally larger for the STP than under the GP for this set of samples.
These effects combine to make it much more likely to see a large outlier close to existing samples under an STP than a GP. As a result, if expected improvement is used to choose the next input to evaluate, an STP is more likely to choose an input close to an existing evaluation, while a GP is more likely to choose a location far from existing samples.

\subsection{Marginal likelihood}

It is often the case in Bayesian optimization that the kernel function contains \emph{hyperparameters} that need to be set, for example the bandwidth in \eqref{eq:kerisose}. One common approach to set those hyperparameters is to maximize the logarithm of the \emph{marginal likelihood} of the data. The marginal likelihood of the STP can be found analytically from the data 
\begin{align}
\label{eq:tmarglike}
    log(p(y|x)) = \gamma((\nu+d)/2) - \gamma(\nu/2) - \frac{d}{2}log(\nu  \pi) - \frac{1}{2}log|\Sigma| - \frac{\nu + d}{2} log \left( 1 + \frac{y^T \Sigma^{-1} y}{\nu}  \right)
\end{align}
where $\gamma$ is the log of the $\Gamma$-function, and again, $\Sigma = \frac{\nu-2}{\nu}K$. The value of $\Sigma$, and so this marginal likelihood, depends on the hyperparameters of the kernel function. So under this maximal marginal likelihood approach,
those hyperparameters are chosen to maximize \eqref{eq:tmarglike}. This procedure could also be used to set $\nu$, though care should be taken to constrain $\nu$. The Student's-T distribution only has finite variance with $\nu > 2$, and  finite kurtosis when $\nu > 4$. While outliers are likely in aerospace, typically our uncertainty is not so large as to have infinite kurtosis, and this knowledge should be encoded into (constraints on) the prior.

\section{Numerical Experiments}

We now compare Student's-T processes with Gaussian processes on problems in Bayesian optimization. We first compare their performances on synthetic benchmark functions commonly used in optimization, and then we compare them on an aerostructural optimization testbed problem. Rather than set $\nu$ via marginal likelihood (as discussed above), we instead use two different STPs, each with a fixed value of $\nu$, to better compare the processes.

\subsection{Synthetic functions}
The first synthetic function is the Rosenbrock function \cite{molga2005test}, given by
\begin{align}
    \nonumber
    f(x_1, x_2) &= (1-x_1)^2 + 100(x_2-x_1^2)^2 \\
    &x_1, x_2 \in [-3,3]
\end{align}
The Rosenbrock function has a single local minimum of $f(x) = 0$ at $x = 0$. The second test function is the six-hump camel function \cite{molga2005test}, given by 
\begin{align}
    \nonumber
    f(x_1, x_2) &= (4 - 2.1x_1^2 + \frac{x_1^4}{3})x_1^2 + x_1x_2 + (-4 + 4x_2^2)x_2^2 \\
    &x_1 \in [-3,3],~x_2 \in [-2,2]
\end{align}
The six-hump camel function has six local minima. Two of these minima are global minima, which have a value of $-1.0316$ at $x = (0.0898,-0.7162)$ and $x = (-0.0898, 0.7162)\}$.

We compare the performance of a Gaussian Process, a Student's-T process with $\nu = 5$, and a Student's-T process with $\nu = 11$, representing the small $\nu$ regime (though still with finite kurtosis) and the medium $\nu$ regime respectively. All processes were set to have a mean function of 0, and to use the squared isotropic kernel function. The Bayesian optimization procedure was carried out identically for all of the processes. First, an initial set of 20 input locations were generated using Latin hypercube sampling, and the objective function was evaluated at each location. The input and output data are scaled to have a mean of 0 and a variance of 1 in each dimension. Using this data, the best value of the kernel bandwidth parameter is found using a two-step grid search. Specifically, the marginal likelihood of the data as a function of the log of the bandwidth parameter is computed for $11$ evenly spaced locations in $[-3, 3]$. The maximum of this initial grid search is used to refine a second grid search, again using $11$ evenly spaced values. Then, the optimization procedure is run for $100$ steps. At each step, the next input evaluated is chosen as the one with the greatest expected improvement. This location is found by first searching a grid with $101$ locations in each dimension, and by using the grid location as the initial location for a local optimization. At every $10$ steps, the evaluated inputs and outputs are re-normalized, and a new optimal bandwidth is found using the previously described procedure. The optimization is run until either $100$ function evaluations have occurred, or until the global optimum is found to within $10^{-4}$. The entire optimization procedure was repeated $100$ different times (with different initial samples) to find the average performance of the processes on the particular optimization problem. Note that for each individual optimization run, all processes begin with the same initial set of function evaluations.

\begin{figure*}
    \centering
    \begin{minipage}[b]{.49\textwidth}
    \includegraphics[width=1.0\textwidth]{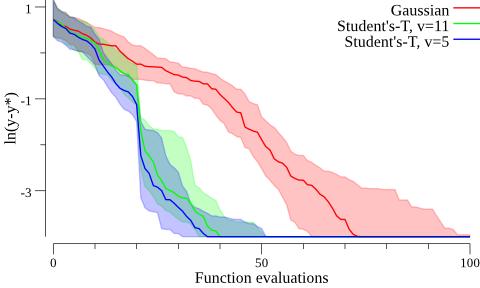}
    \caption{A comparison of the GP against two STPs with different values of $\nu$ for the Rosenbrock test case. The y-axis shows $log_{10} (\hat{y}-y^*)$ as a function of the optimization step. The solid line shows the median performance, and the shaded region covers the inner quartiles. Both STPs significantly outperform the GP.}\label{fig:rosenresults}
    \end{minipage}\hfill
    \begin{minipage}[b]{.49\textwidth}
    \includegraphics[width=1.0\textwidth]{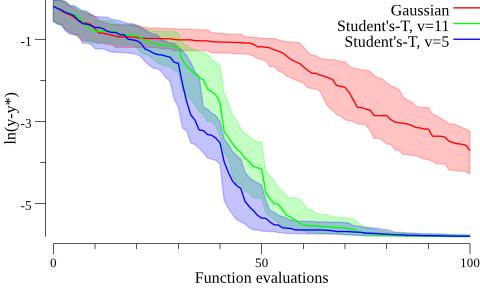}
    \caption{A comparison of the GP against two STPs with different values of $\nu$ for the six-hump camel test case. The y-axis shows $log_{10} (\hat{y}-y^*)$ as a function of the optimization step.  The solid line shows the median performance, and the shaded region covers the inner quartiles. Both STPs significantly outperform the GP.}\label{fig:camelresults}
    \end{minipage}
\end{figure*}
Fig. \ref{fig:rosenresults} compares the performance on the Rosenbrock test case for the three processes. The $y$-axis shows the log of the difference between the current best found value and the global optimum $y^*$, with difference capped at a minimum of $10^{-4}$. The $x$ axis shows how the average of this value changes as a function of the optimization step. The shaded region shows the inner quartiles of performance over the $100$ runs, and the solid line depicts the median performance. Similarly, Fig. \ref{fig:camelresults} shows the performance on the six-hump camel test case. It can be seen that while all processes find a significantly better value than the initial samples, the both of the STPs significantly outperform the GP. The STPs both find a better final value, and hone in on the optimum earlier in the optimization. Note also that the STPs are significantly more robust. Nearly all of the runs of the STP runs find the global minimum within the allotted budget, while in many cases the GP fails to find the best input. This is especially true for the six-hump camel function which contains many local minima, exactly the case where Bayesian optimization is most appropriate. The STP with $\nu = 5$ seems to outperform the STP with $\nu = 11$, though the differences are more minor compared with the differences with the GP.

\subsection{Aerostructural optimization}

\begin{figure*}
    \centering
    \begin{minipage}[b]{.46\textwidth}
    \includegraphics[width=1.0\textwidth]{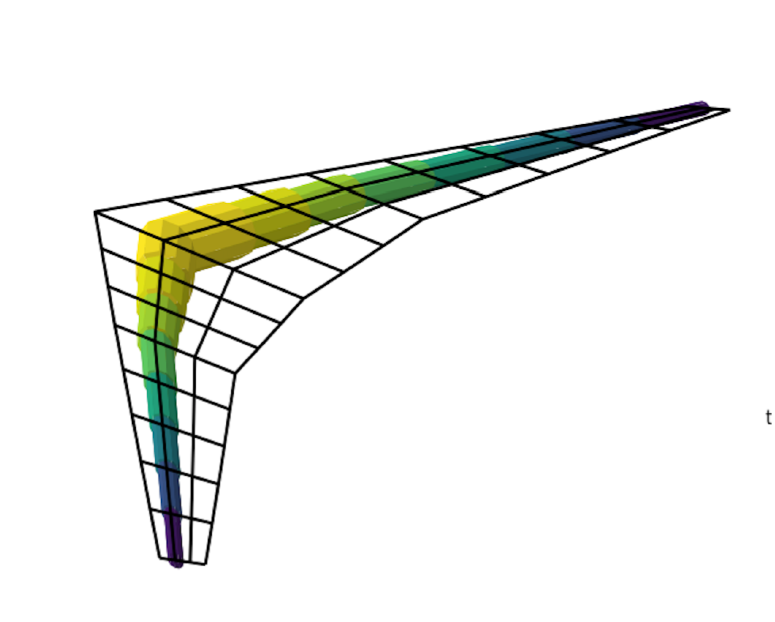}
    \caption{A depiction of the OpenAeroStruct problem containing wing planform and internal structure}\label{fig:aerostructfig}
    \end{minipage}\hfill
    \begin{minipage}[b]{.46\textwidth}
    \includegraphics[width=1.0\textwidth]{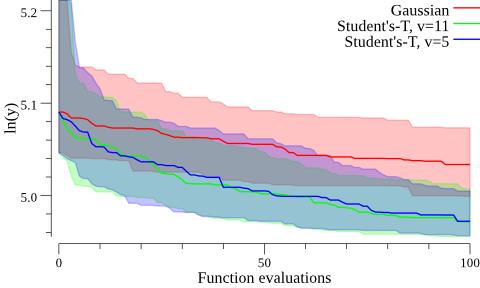}
    \caption{A comparison of the GP against two STPs with different values of $\nu$ for the OpenAeroStruct test case. The y-axis shows $log_{10}(\hat{y})$ as a function of the optimization step. The solid line shows the median performance, and the shaded region covers the inner quartiles. The STPs outperform the GP, though the two STPs behave nearly identically.}\label{fig:openaeroresults}
    \end{minipage}
\end{figure*}

The last test case is to find the optimal wing design for a coupled aerostructural problem, shown in Fig. \ref{fig:aerostructfig}. The coupled solver code is the OpenAeroStruct package \cite{Jasa2018}, which is implemented on top of the OpenMDAO framework \cite{gray2010openmdao}. This test case has a 7-dimensional input, a univariate output objective, and two non-linear constraints. The finput parameters are a) the angle of attack of the wing, constrained to be in $[-10,10]$ degrees b) three wing thickness parameters constrained to be in $[0.01, 0.5]$ c) three wing twist parameters, constrained to be in $[-15,15]$ degrees. The objective is to minimize the fuel burn of a simulated mission, and the the wing twist and thickness are constrained so that a) the design is physically realizable (the structure does not intersect with itself) and b) the aerodynamic forces do not cause the wing to break. We transform this constrained optimization problem into the following unconstrained problem for demonstration purposes. If the constraints are not violated, the objective is simply taken as the fuel burn. If the constraints are violated, then then the objective is taken to be $180000(1+ \xi)$, where $\xi$ is the constraint violation (the value of $180000$ was chosen to be larger than the fuel burn of a valid design) \footnote{For some input conditions, the design constraints are satisfied, but a negative fuel burn is returned. This is clearly non-physical, and in this case the objective value is taken to be $300000$. There are a small number of inputs for which the simulation crashes, and in this case the objective is taken to be $400000$.}. The optimization is performed almost identically as described above for the synthetic functions, except in a $7$ dimensional input space a full grid search for the best EI is too computationally intensive. Instead, $10,000$ test candidates are generated from Latin-hypercube sampling, and the best of those locations is used as the input to the local optimization step.

Fig. \ref{fig:openaeroresults} compares the performance of the GP against the two STPs for the aerostructural design problem. Here, the plot shows the log of the current best value (rather than the normalized best value) since the global optimum is unknown. While the difference isn't as stark as the analytic cases, it is still clear that the STPs outperform the GP on this test case. The median performance is consistently better as the optimization progresses, and as the optimization continues, the $75^{th}$ percentile of the STPs is close to outperforming the $25^{th}$ percentile of the GP. The difference between the performance between the two STPs is insignificant for this problem.

\section{Conclusion}

In this paper we began by presenting the Student's-T process, a prior over functions based on the multivariate Student's-T distribution. The STP has similar desirable properties to 
those of a Gaussian process, in that it has a simple expression for marginal distributions, and it has an analytic Bayesian update rule when new samples are observed. The STP also has some significant advantages over a Gaussian process. First, outliers are likely to occur under an STP with a small value of the degrees of freedom parameter. Second, the posterior covariance adjusts depending on the actual function values observed (and not just their locations), increasing if the samples vary by more than expected, or decreasing if they vary by less than expected. We
then presented an analytic expression for the expected improvement under a Student's-T distribution, and showed how to set kernel hyperparameters using marginal likelihood. Finally, we presented numerical simulation results that show STPs outperform GPs on several synthetic benchmarks as well as an aerostructural design optimization problem.

It is known that STPs cannot be better than GPs for \emph{every} possible design problem \cite{wolpert1997no}. However, it seems that the STP prior may be more naturally suited to problems in aerospace optimization, which do often have large outliers and the need for adjusted variance. With these advantages, and no obvious disadvantages, it seems natural to ``upgrade'' from Gaussian processes to Student's-T processes in many aerospace design applications. Future work remains to extend Student's-T processes to other important ideas from Bayesian optimization, such as extending Student's-T processes to constrained optimization and multi-fidelity optimization, and also to other search strategies, such as the Knowledge Gradient algorithm.

\section{Acknowledgements}
This work was made possible through the support of AFOSR MURI on multi-information
sources of multi-physics systems under Award Number FA9550-15-1-0038. We would also like to thank the Santa Fe Institute for support of this research.

\bibliography{studenttrefs}

\end{document}